\documentclass[conference]{IEEEtran}
\IEEEoverridecommandlockouts
\usepackage{iftex}
\ifXeTeX
  \usepackage{fontspec}
\fi

\usepackage{cite}
\usepackage{amsmath,amssymb,amsfonts}
\usepackage{graphicx}
\usepackage{textcomp}
\usepackage{xcolor}
\usepackage{booktabs}
\usepackage{multirow}
\usepackage{siunitx}
\usepackage{pgfplots}
\pgfplotsset{compat=1.17}
\usepackage[hidelinks]{hyperref}

\definecolor{seriesA}{HTML}{2A78D6}  
\definecolor{seriesB}{HTML}{EB6834}  
\definecolor{seriesC}{HTML}{1BAF7A}  
\definecolor{seriesD}{HTML}{4A3AA7}  
\definecolor{gridgray}{HTML}{D9D9D6}

\pgfplotsset{
  rdstyle/.style={
    width=\columnwidth, height=0.72\columnwidth,
    log basis x=2,
    xmin=0.8, xmax=40,
    xtick={1,2,4,8,16,32}, xticklabels={1,2,4,8,16,32},
    xlabel={Bits per complex sample (log scale)},
    grid=major, major grid style={gridgray, thin},
    axis line style={black!60}, tick style={black!60},
    label style={font=\footnotesize}, tick label style={font=\footnotesize},
    legend style={font=\scriptsize, draw=black!30, fill=white, fill opacity=0.9, text opacity=1},
    legend cell align=left,
  },
}
\tikzset{
  aeline/.style={seriesA, line width=1pt, mark=*, mark size=2pt, solid},
  baqline/.style={seriesB, line width=1pt, mark=square*, mark size=2pt, dashed},
  baqtline/.style={seriesB, line width=1pt, mark=square, mark size=2.2pt, solid, mark options={solid, fill=white}},
  fftline/.style={seriesC, line width=1pt, mark=triangle*, mark size=2.6pt, densely dotted},
  kltline/.style={seriesD, line width=1pt, mark=diamond*, mark size=2.6pt, dashdotted},
  bkltline/.style={seriesD, line width=1pt, mark=diamond, mark size=2.8pt, loosely dashed, mark options={solid, fill=white}},
}

\begin{document}

\title{Learned Compression of SAR Phase-History\\
Data: A Rate-Honest Feasibility Study\\
on GOTCHA}

\author{\IEEEauthorblockN{Alizishaan Khatri}
\IEEEauthorblockA{Wrynx Inc\\
research@wrynx.com}}

\maketitle

\begin{abstract}
On-board compression of synthetic aperture radar (SAR) phase history is
bandwidth-critical, and block-adaptive quantization (BAQ) remains the
operational standard. We test whether a small convolutional autoencoder, with
its encoder on the sensor, can compete with BAQ on complex phase-history
patches from the AFRL GOTCHA collection. Every method is charged for all
transmitted bits, rates are reported in bits per complex sample (b/cs), and
detection is scored by one-to-one matching of CA-CFAR detections. The
autoencoder (\num{28656} encoder parameters) loses at every rate. At 16~b/cs
it reaches \SI{-2.87}{dB} NMSE, against \SI{-35.5}{dB} for 8-bit BAQ with
$\pm3\sigma$ clipping and \SI{-41.0}{dB} with a tuned clipping range. It also
loses to a $16\times16$ block Karhunen--Lo\`eve transform (KLT), a local
linear coder with a tenth of its encoder cost (\SI{-5.39}{dB}). Running the
network in a companded Fourier domain helps, but its detection F1 remains
bounded at 33\%. The evidence points to this model, its normalization, and its
objective, not to a fundamental limit of learned coding. Per patch, the data
have modest lag-1 coherence ($|\rho|\approx0.3$) and patch-specific spectral
concentration. Two findings concern evaluation itself. First, 97\% of CFAR
crossings on raw $64\times64$ patches are border artifacts of the zero-padded
detector. Second, on interior cells BAQ's clipping range decides detection:
8-bit BAQ keeps 69\% F1 with tuned clipping but 17\% at $\pm3\sigma$, and at
8~b/cs or less adaptive FFT thresholding preserves more detections than BAQ.
We close with an evaluation protocol for learned radar compression.
\end{abstract}


\section{Introduction}

SAR sensors produce raw data faster than their downlinks can carry it, so the
data are compressed on board before image formation, within tight power and
latency budgets. The operational standard is block-adaptive quantization
(BAQ)~\cite{kwok1989baq}, which normalizes each block of I/Q samples by its
local statistics and quantizes it to a few bits. Sentinel-1 uses a flexible,
dynamically allocated variant (FDBAQ)~\cite{attema2010fdbaq}. BAQ is cheap and
robust, and it is hard to beat on raw data~\cite{benz1995comparison}.

Learned transform coding has overtaken classical codecs on natural
images~\cite{balle2017end,balle2018variational,theis2017lossy}. It is natural
to ask whether a learned encoder small enough for a sensor platform can
compress radar phase history better than BAQ. This paper reports a
feasibility study on that question using the AFRL GOTCHA
dataset~\cite{ertin2007gotcha,casteel2007challenge}. Several easy-to-make
evaluation errors each made the learned coder look better than it is; we
distill them into a protocol (Section~\ref{sec:protocol}). Our contributions
are:
\begin{enumerate}
  \item A \emph{compressibility characterization} of GOTCHA phase-history
        patches on the full test set.
  \item A \emph{split autoencoder} with the encoder on the sensor and the
        decoder on the ground, and a variant in a companded Fourier domain.
  \item A \emph{rate-honest benchmark} in bits per complex sample. It includes
        BAQ with default and tuned clipping and the entropy of its indices,
        FFT hard-thresholding, global and $16\times16$ block KLT coders, and an
        FP16 cast, with every side-information bit charged. Detection is
        scored one-to-one on interior cells, away from detector border
        artifacts.
  \item A \emph{scoped negative result} and an \emph{evaluation protocol}.
\end{enumerate}

\section{Related Work}

\emph{Raw SAR data compression.} BAQ was developed for
Magellan~\cite{kwok1989baq}, and its descendants remain standard. Benz
\emph{et al.}~\cite{benz1995comparison} compared BAQ with vector quantization
and transform coding of raw data. FDBAQ~\cite{attema2010fdbaq} adapts the
quantizer to the signal-to-clutter ratio. Entropy-constrained BAQ adds
variable-length coding of the quantizer indices~\cite{algra2002ecbaq}. A
learned coder that is itself entropy coded should be compared with an
entropy-coded BAQ.

\emph{Transform coding.} For Gaussian sources under MSE, the KLT followed by
scalar quantization is the classical optimum among fixed linear
transforms~\cite{goyal2001transform}, and Lloyd--Max
quantizers~\cite{max1960quantizing} optimize the scalar step. A linear
autoencoder trained with MSE converges to the principal subspace, that is, to
the KLT~\cite{baldi1989pca}.

\emph{Learned compression.} End-to-end optimized
codecs~\cite{balle2017end,theis2017lossy} and hyperprior entropy
models~\cite{balle2018variational} jointly train a transform, a quantizer, and
an entropy model under a rate-distortion objective. The models studied here
are not trained that way (Section~\ref{sec:method}).

\emph{Deep learning on complex SAR data.} Complex-valued
networks~\cite{trabelsi2018deep} and coherence-preserving
autoencoders~\cite{asiyabi2023complex} have been applied to focused
single-look complex imagery. That imagery has spatial structure that phase
history lacks, so their figures are not directly comparable to ours.

\section{Data and Characterization}
\label{sec:data}

\subsection{What ``Raw'' Means Here}
\label{sec:raw}

The public GOTCHA release holds 8 circular passes, 4 polarizations, and 360
files per pass and polarization, one per degree of azimuth, for \num{11520}
files~\cite{casteel2007challenge}. Each file stores the phase history as a
complex single-precision array of 424 frequency samples
(\SIrange{9.288}{9.910}{GHz}, \SI{1.47}{MHz} spacing) by about 117 pulses,
with the per-pulse range to scene center. These are k-space samples referenced
to the scene center. They are not the analog-to-digital converter (ADC) stream
that an on-board BAQ processes. Two consequences follow. First, the 2-D FFT of
a patch is approximately a coarse sub-aperture image, so Fourier-domain
structure (Section~\ref{sec:char}) largely reflects bright scatterers in the
scene. Second, rates quoted against FP32 storage overstate every method: a
real front end produces 8--12~bits per I/Q component, or 16--24~bits per
complex sample. We therefore report rates in \emph{bits per complex sample}
(b/cs). The FP32 compression ratio $\mathrm{CR}=64/b$ is given only for
reference.

\subsection{Tiling, Split, and Normalization}

Each file's array is tiled into non-overlapping $64\times64$ patches from its
first sample, giving 6 patches per file. Tiling discards the last 40 of 424
frequency samples (9\%) and about 53 of 117 pulses (45\%). Patch rows index
frequency and columns index pulse (slow time). Each patch therefore spans only
\SI{94}{MHz} of the \SI{622}{MHz} band and 64 pulses (about \ang{0.55} of
azimuth), which limits the structure any coder can exploit within one patch.
Each patch is stored as two real channels (I and Q), for \num{69120} patches in
total.

The split is random \emph{at the file level} (seed 42): \num{9216} training,
\num{1152} validation, and \num{1152} test files, or \num{55296}, \num{6912},
and \num{6912} patches. Because polarizations of the same pass and azimuth
are separate files, every test file shares its pass and azimuth degree with
at least one training file. HV and VH are nearly identical by reciprocity, and
HH and VV are strongly correlated, so many test patches have near-duplicates
in training. This can only flatter the learned methods. BAQ, FFT
thresholding, and the FP16 cast learn nothing and are unaffected. For the two
KLT coders we measured the effect directly. Refitting each with the test
patch's whole pass held out changes NMSE by \SI{0.16}{dB} (global KLT) and
less than \SI{0.01}{dB} (block KLT). The autoencoders were not retrained on a
held-out split.

All data are divided by one global max-abs value computed on the training set
($1.47\times10^{-2}$). After this scaling, the median sample magnitude is
0.030. Patch RMS levels span \SIrange{-35.7}{-19.8}{dB} relative to full scale
(5th to 95th percentile), a \SI{16}{dB} spread. Within a patch, the $16\times16$
block standard deviations differ by only \SI{1.9}{dB} (median max/min ratio).
The per-block gain that BAQ estimates is therefore mostly a per-patch gain.

\subsection{Compressibility Characterization}
\label{sec:char}

We measure the lag-1 complex coherence along each axis,
\begin{equation}
  \rho = \frac{\bigl|\mathbb{E}\bigl[s_{n}\,s_{n+1}^{*}\bigr]\bigr|}
              {\sqrt{\mathbb{E}|s_n|^2\,\mathbb{E}|s_{n+1}|^2}},
\end{equation}
with expectations taken within each patch and $|\rho|$ then averaged over
patches. For comparison we also give the \emph{pooled} estimate, with
expectations taken over all patches at once. We also measure the fraction of
energy in the top $q$ of 2-D FFT bins, which for white complex Gaussian data
is $q(1+\ln(1/q))$, and the energy captured by the leading half of the KLT
modes (Section~\ref{sec:baselines}). Table~\ref{tab:char} lists all
quantities.

\begin{table}[t]
\centering
\caption{Compressibility of GOTCHA phase-history patches}
\label{tab:char}
\setlength{\tabcolsep}{4pt}
\begin{tabular}{lccc}
\toprule
Quantity (test set unless noted) & Freq. & Pulse & White \\
\midrule
$|\rho|$, per patch (mean)            & 0.268 & 0.313 & 0.014 \\
$|\rho|$, pooled over all patches     & 0.021 & 0.126 & $\approx0$ \\
\midrule
Energy in top 1\% of FFT bins         & \multicolumn{2}{c}{0.372} & 0.056 \\
Energy in top 10\% of FFT bins        & \multicolumn{2}{c}{0.671} & 0.330 \\
\midrule
Top 2048 of 4096 KLT modes: & & & \\
\quad eigenvalue share (training set) & \multicolumn{2}{c}{0.813} & 0.500 \\
\quad energy captured, pooled         & \multicolumn{2}{c}{0.733} & 0.500 \\
\quad energy captured, per-patch mean & \multicolumn{2}{c}{0.664} & 0.500 \\
\bottomrule
\end{tabular}
\end{table}

Per patch, $|\rho|\approx0.3$ on both axes, 20 times the white-noise value.
Each patch's lag-1 correlation has a phase set by its scatterers' positions,
so the pooled estimate cancels it and can be misleadingly small. The
correlation is still weak for predictive coding: a first-order predictor
gains only $-10\log_{10}(1-|\rho|^2)\approx$ \SIrange{0.3}{0.45}{dB}. The
Fourier concentration is strong (the top 1\% of bins hold $6.6\times$ the
white-noise share), but the energetic bins differ from patch to patch.
Per-patch adaptive selection captures 67\% of the energy with 10\% of the
bins, whereas a single global KLT captures 73\% of test energy with half of
its modes. The KLT captures less on test data than its training eigenvalue
share (81\%). Its per-patch mean (66\%) is lower still, because the
high-energy, scatterer-dominated patches that dominate the pooled figure are
the ones it captures best. The pooled test figure is the one that sets NMSE:
$10\log_{10}(1-0.733)=\SI{-5.73}{dB}$, the value in Table~\ref{tab:rd}.

\section{Method}
\label{sec:method}

\subsection{Split Autoencoder Architecture}

The encoder runs on the sensor and the decoder on the ground.
Table~\ref{tab:arch} lists the layers. The encoder downsamples twice with
strided $3\times3$ convolutions, each followed by batch
normalization~\cite{ioffe2015batch}, GELU~\cite{hendrycks2016gelu}, and a
residual block~\cite{he2016deep}. This yields a real latent
$\mathbf{z}\in\mathbb{R}^{C\times16\times16}$, and each latent value has a
$31\times31$-sample receptive field. The decoder mirrors the encoder with
$4\times4$ stride-2 transposed convolutions. The latent is sent as FP16 with
no entropy coding, so the rate is
\begin{equation}
  b = \frac{16\cdot C\cdot 16\cdot 16}{64\cdot64} = C\ \text{b/cs},
  \label{eq:cr}
\end{equation}
and $C\in\{4,8,16,32,64\}$ gives 4--64~b/cs. Whether the latents tolerate
coarser quantization, and how much entropy coding would save, was not tested.

\begin{table}[t]
\centering
\caption{Autoencoder layers (latent width $C$; output $C_\text{out}\times H\times W$)}
\label{tab:arch}
\begin{tabular}{lll}
\toprule
 & Layer & Output \\
\midrule
\multirow{4}{*}{\rotatebox{90}{Encoder}}
 & Conv $3\times3$/2, BN, GELU & $32\times32\times32$ \\
 & ResBlock(32)                & $32\times32\times32$ \\
 & Conv $3\times3$/2, BN, GELU & $C\times16\times16$ \\
 & ResBlock($C$)               & $C\times16\times16$ \\
\midrule
\multirow{4}{*}{\rotatebox{90}{Decoder}}
 & ResBlock($C$)                    & $C\times16\times16$ \\
 & ConvT $4\times4$/2, BN, GELU     & $32\times32\times32$ \\
 & ResBlock(32)                     & $32\times32\times32$ \\
 & ConvT $4\times4$/2               & $2\times64\times64$ \\
\bottomrule
\multicolumn{3}{l}{\footnotesize ResBlock: $x + \mathrm{BN}(\mathrm{Conv}(\mathrm{GELU}(\mathrm{BN}(\mathrm{Conv}(x)))))$, $3\times3$ kernels.}
\end{tabular}
\end{table}

\subsection{Target-Focal Coherent Loss}

Plain MSE collapses to a zero output (Section~\ref{sec:dev}). We instead
combine an $\ell_1$ term, an amplitude-weighted magnitude term, and an
amplitude-weighted phase term. Let $s=x_I+jx_Q$ and $\hat s$ be input and
reconstruction, with $a=|s|/\max|s|$ per patch:
\begin{multline}
  \mathcal{L} = \|\hat{\mathbf{x}}-\mathbf{x}\|_1
  + \overline{(1+\alpha a^{\gamma})\,(|\hat s|-|s|)^2} \\
  + \beta\,\overline{a^{\gamma}\,\bigl(1-\cos(\angle(\hat s+\epsilon)-\angle(s+\epsilon))\bigr)},
  \label{eq:loss}
\end{multline}
where the overbar is a mean over samples, $\alpha=5$, $\gamma=2$, $\beta=1$,
and $\epsilon=10^{-6}$. The focal weight~\cite{lin2017focal} emphasizes strong
scatterers. The $\epsilon$ offset avoids NaN gradients of $\angle(\cdot)$ at
the origin. After normalization, $\epsilon$ is $3\times10^{-5}$ of the median
$|s|$, and only 0.003\% of samples lie below $10^{-5}$, so it does not bias
the phase of typical samples. The loss is not MSE, so the NMSE comparisons
below are not what the network was trained to minimize.

\subsection{Transform-Domain Variant}

A variant computes $S=\mathrm{fftshift}(\mathrm{FFT2}(s))$ with orthonormal
scaling and applies an invertible magnitude compander
\begin{equation}
\begin{aligned}
  \mathcal{C}(S) &= \frac{S}{|S|}\,\log\!\left(1+\frac{|S|}{\sigma}\right),\\
  \mathcal{C}^{-1}(Z) &= \frac{Z}{|Z|}\,\sigma\left(e^{|Z|}-1\right),
\end{aligned}
\end{equation}
with $\sigma=0.0213$, the median training spectral magnitude. The same network
runs on the companded spectrum. Its output is expanded and inverse
transformed, so the loss and metrics still apply to I/Q samples, at the same
rate as the sample-domain model.

\subsection{Training}

Models are trained for 15 epochs with AdamW~\cite{loshchilov2019decoupled}
(learning rate $5\times10^{-4}$, weight decay $10^{-4}$), batch size 64, and
gradient-norm clipping at 1.0. Batches with a non-finite loss are skipped. We
keep the checkpoint with the best validation loss.

\subsection{Baselines and Rate Accounting}
\label{sec:baselines}

\emph{FP16 cast.} Each I and Q value is rounded to FP16 (32~b/cs), a
reference point for the $C=32$ autoencoder.

\emph{BAQ.} Each $16\times16$ block of each I/Q channel is normalized by its
standard deviation and quantized with an $n$-bit mid-rise uniform quantizer
over $\pm c\sigma$. The rate is $2n$~b/cs plus one FP16 scale per block and
channel ($2n+0.125$~b/cs). We report $c=3$ and, per bit depth, the $c$ that
minimizes validation NMSE (grid 1--8 in steps of 0.25). We also report $H$,
the zeroth-order entropy of the quantizer indices plus the 0.125~b/cs of
scales. This is the rate an ideal memoryless entropy coder would need, a
proxy for entropy-constrained BAQ~\cite{algra2002ecbaq}.

\emph{FFT hard-thresholding.} Per patch, the largest fraction $k$ of 2-D FFT
coefficients is kept at 32 bits each (FP16 complex), plus a 4096-bit position
map: $32k+1$~b/cs.

\emph{Global KLT.} We estimate the $4096\times4096$ complex covariance of
vectorized training patches, keep its top $K$ eigenvectors, and send the $K$
coefficients as FP16 complex values with no side information: $K/128$~b/cs.
Its encoder is a patch-wide projection with $4096K$ complex weights
(\SI{67}{MB} at $K=2048$) and $4K$ real MAC per complex sample (8192 at
$K=2048$). No $31\times31$ convolutional encoder can represent it, and it is
not deployable. We use it as an upper-bound anchor for fixed linear coding.

\emph{Block KLT.} The like-for-like linear baseline for a local
convolutional encoder is a $16\times16$ block KLT: the same procedure on the
256-dimensional vectors of $16\times16$ blocks, keeping $k$ of 256
coefficients per block ($k/8$~b/cs). Its encoder has $256k$ complex weights
(\SI{256}{KB} at $k=128$) and needs $4k$ real MAC per complex sample, 512 at
16~b/cs, about a tenth of the autoencoder's \num{5300}. FP16 coefficients
carry their own exponent, so no explicit block scaling is needed.

\subsection{Metrics}

\emph{NMSE} is $10\log_{10}\bigl(\sum\|\hat{\mathbf{x}}-\mathbf{x}\|^2/
\sum\|\mathbf{x}\|^2\bigr)$ over the full test set. \SI{0}{dB} corresponds to
outputting zeros.

\emph{CFAR detection proxy.} A 2-D CA-CFAR
detector~\cite{finn1968adaptive,richards2014fundamentals} runs on $|s|^2$
and $|\hat s|^2$ per patch, with guard radius 2, training band 4 ($N=144$
cells), design $P_{fa}=10^{-4}$, and threshold multiplier
$N(P_{fa}^{-1/N}-1)$. The local mean is computed by convolution with zero
padding. In the 6-cell border band (34\% of cells), the training window
extends past the patch, the mean is underestimated, and the detector fires
far too often (Section~\ref{sec:cfar}). We therefore score detection on the
\emph{interior} $52\times52$ cells, where the window is complete. We report
the following:
\begin{itemize}
  \item \emph{Recall, precision, and F1} under a maximum one-to-one matching
        of raw and decoded detections within $\pm1$ cell (Chebyshev),
        computed per patch by the Hungarian
        algorithm~\cite{kuhn1955hungarian}. We give them on interior cells and,
        for comparison, on all cells.
  \item \emph{Retention}: decoded detections within $\pm1$ cell of any raw
        detection, divided by the number of raw detections, on all cells. This
        many-to-one score is an upper bound on all-cell recall. It is the only
        detection score available for the autoencoders, whose checkpoints were
        not retained.
  \item The false-alarm shift
        $\Delta P_{fa}=|N_{\text{dec}}-N_{\text{raw}}|/N_{\text{cells}}$.
\end{itemize}
This proxy is not target detection (Section~\ref{sec:cfar}).

\section{Results}
\label{sec:results}

\subsection{Development Trajectory}
\label{sec:dev}

Table~\ref{tab:dev} summarizes the design iterations at 16~b/cs
($C=16$). The first model (V0) had a fully connected 64-dimensional
bottleneck and an MSE loss. It learned to output zeros (NMSE \SI{0.00}{dB}).
Switching to (\ref{eq:loss}), a convolutional latent, and the
$\epsilon$-guard produced non-trivial reconstructions (V2). Runs without the
$\epsilon$-guard, or with a loss on the dB peak-to-background ratio at
CFAR-relevant cells, or with a PixelShuffle
decoder~\cite{shi2016real,odena2016deconvolution}, were no better and are
omitted. V6 is V2 with seeding and best-validation checkpointing. V7 is V6 in
the companded Fourier domain. Because the checkpoints were not retained,
detection is available only as all-cell retention and the decoded detection
count. These give upper bounds: recall is at most the retention, and
precision is at most $\text{retention}\times N_{\text{raw}}/N_{\text{dec}}$.

\begin{table}[t]
\centering
\caption{Design iterations at 16 b/cs, test set, all cells. $N_\text{dec}$:
decoded CFAR detections ($N_\text{raw}=\num{41774}$). F1 is an upper bound.}
\label{tab:dev}
\setlength{\tabcolsep}{3.5pt}
\begin{tabular}{lrrrr}
\toprule
Model & NMSE (dB) & Ret.\ (\%) & $N_\text{dec}$ & F1 (\%) \\
\midrule
V0 FC, MSE$^\dagger$   & 0.00    & --    & --          & -- \\
V2 loss (\ref{eq:loss})& $+0.02$ & 13.87 & \num{17082} & $\le19.7$ \\
V6 seeded, best-val    & $-2.49$ & 14.05 & \num{16940} & $\le20.0$ \\
V6, 3 seeds            & --      & $13.3\pm1.1$ & --   & $\le19.1$ \\
V7 companded FFT       & $\mathbf{-3.86}$ & \textbf{36.08} & \num{48355} & $\le33.5$ \\
\bottomrule
\multicolumn{5}{l}{\footnotesize $^\dagger$Fully connected 64-d bottleneck, 0.25 b/cs.}
\end{tabular}
\end{table}

Three independent training runs of the same configuration and seed (V6, seed
0) gave \SI{-2.49}{dB}/14.05\%, 14.86\% retention, and
\SI{-2.87}{dB}/12.55\% (the run used in Table~\ref{tab:rd}). They differ only
in nondeterministic GPU kernels and in how much of the random-number stream
was consumed before training. Run-to-run spread is thus about \SI{0.4}{dB}
and 2.3 retention points. V7 roughly triples retention but produces
\emph{more} detections than the raw data, and its precision bound (31\%) is no
better than V6's (35\%). Its F1 bound therefore rises only from 20\% to 33\%.
These all-cell scores are dominated by border cells (Section~\ref{sec:cfar}).

\subsection{Rate-Distortion Comparison}
\label{sec:rd}

\begin{figure}[t]
\centering
\begin{tikzpicture}
\begin{axis}[xmode=log, rdstyle,
  ylabel={NMSE (dB), lower is better},
  ymin=-44, ymax=2,
  legend pos=south west]
\addplot[aeline]  coordinates {(4,-1.06) (8,-1.88) (16,-2.87) (32,-4.73)};
\addlegendentry{Autoencoder}
\addplot[baqline] coordinates {(2.125,-0.73) (4.125,-7.23) (6.125,-13.21) (8.125,-19.14) (12.125,-29.96) (16.125,-35.47)};
\addlegendentry{BAQ, $\pm3\sigma$}
\addplot[baqtline] coordinates {(2.125,-4.46) (4.125,-9.41) (6.125,-14.48) (8.125,-19.60) (12.125,-30.19) (16.125,-41.01)};
\addlegendentry{BAQ, tuned clip}
\addplot[fftline] coordinates {(1.32,-2.63) (1.64,-3.33) (2.6,-4.58) (4.2,-5.89) (7.4,-7.83) (13.8,-11.37)};
\addlegendentry{FFT thresholding}
\addplot[kltline] coordinates {(1,-1.03) (2,-1.54) (4,-2.22) (8,-3.34) (16,-5.73)};
\addlegendentry{Global KLT}
\addplot[bkltline] coordinates {(1,-0.55) (2,-1.02) (4,-1.71) (8,-2.92) (16,-5.39)};
\addlegendentry{$16\times16$ block KLT}
\end{axis}
\end{tikzpicture}

\vspace{2mm}
\begin{tikzpicture}
\begin{axis}[xmode=log, rdstyle,
  ylabel={Interior CFAR F1 (\%)},
  ymin=-3, ymax=75]
\addplot[baqline] coordinates {(2.125,0) (4.125,0) (6.125,0.16) (8.125,4.51) (12.125,13.54) (16.125,16.67)};
\addplot[baqtline] coordinates {(2.125,0) (4.125,0) (6.125,0) (8.125,0) (12.125,31.89) (16.125,68.72)};
\addplot[fftline] coordinates {(1.32,1.07) (1.64,2.81) (2.6,7.66) (4.2,15.82) (7.4,26.00) (13.8,44.82)};
\addplot[kltline] coordinates {(1,0) (2,0) (4,0.50) (8,1.46) (16,10.91)};
\addplot[bkltline] coordinates {(1,0) (2,0) (4,0.28) (8,1.23) (16,9.01)};
\end{axis}
\end{tikzpicture}
\caption{Rate-distortion on the GOTCHA test set, with rates in bits per
complex sample including all side information (FP32 storage is 64~b/cs).
Top: NMSE. Bottom: one-to-one CFAR F1 on interior cells, with line styles as
in the top panel. The autoencoder is absent from the bottom panel because its
checkpoints were not retained; its all-cell retention is in
Table~\ref{tab:rd}. FP16 cast (not plotted): \SI{-73.7}{dB} and 99.96\%
interior F1 at 32~b/cs.}
\label{fig:rd}
\end{figure}
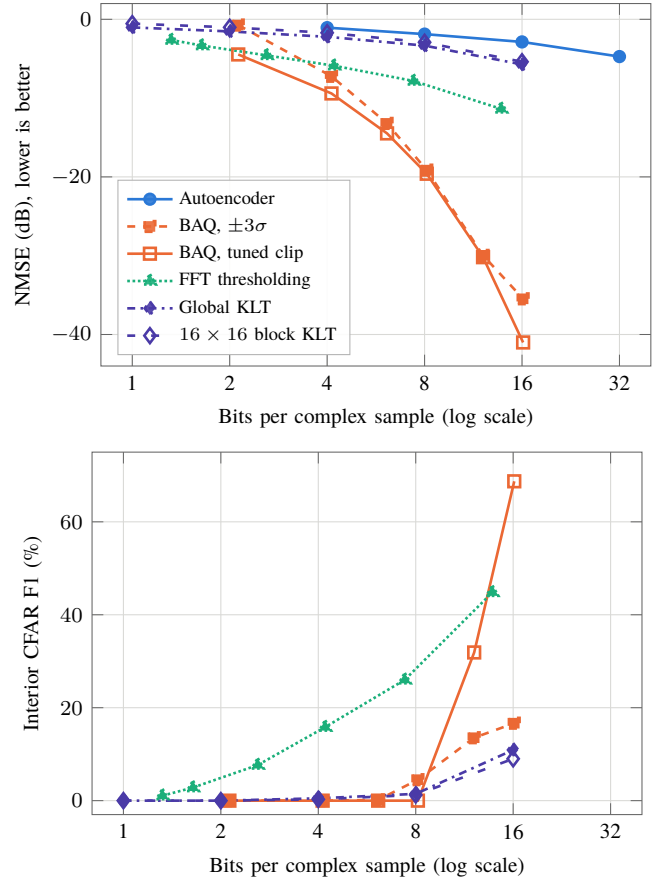

Fig.~\ref{fig:rd} and Table~\ref{tab:rd} compare the autoencoder sweep (seed
0, best-val) with all baselines.

\begin{table*}[t]
\centering
\caption{Rate-distortion and detection on the full test set. b/cs: bits per complex sample (FP32 = 64). Ret.: many-to-one retention, all cells.
F1 and $\Delta P_{fa}$: one-to-one, all cells. R/P/F1$_\text{int}$: one-to-one on interior cells (\num{1250} raw detections).
$H$: BAQ rate with ideal entropy coding of indices, including scales.}
\label{tab:rd}
\setlength{\tabcolsep}{4.6pt}
\begin{tabular}{lrrrrrrrrrr}
\toprule
 & & & & \multicolumn{3}{c}{All cells} & \multicolumn{3}{c}{Interior cells} & \\
\cmidrule(lr){5-7}\cmidrule(lr){8-10}
Method & b/cs & CR & NMSE (dB) & Ret.\ (\%) & F1 (\%) & $\Delta P_{fa}$ ($10^{-4}$) & R (\%) & P (\%) & F1 (\%) & $H$ (b/cs) \\
\midrule
FP16 cast                         & 32.0  & 2.0  & $-73.67$ & 99.96 & 99.94 & 0.00 & 100.00 & 99.92 & 99.96 & -- \\
\midrule
AE, $C=32$                        & 32.0  & 2.0  & $-4.73$  & 40.35 & -- & -- & -- & -- & -- & -- \\
AE, $C=16$                        & 16.0  & 4.0  & $-2.87$  & 12.55 & -- & -- & -- & -- & -- & -- \\
AE, $C=8$                         & 8.0   & 8.0  & $-1.88$  & 7.19  & -- & -- & -- & -- & -- & -- \\
AE, $C=4$                         & 4.0   & 16.0 & $-1.06$  & 6.18  & -- & -- & -- & -- & -- & -- \\
\midrule
BAQ 8-bit, $\pm3\sigma$           & 16.1  & 4.0  & $-35.47$ & 78.52 & 87.21 & 3.0  & 9.12  & 96.61 & 16.67 & 15.02 \\
BAQ 6-bit, $\pm3\sigma$           & 12.1  & 5.3  & $-29.96$ & 75.04 & 83.95 & 3.3  & 7.28  & 96.81 & 13.54 & 11.03 \\
BAQ 4-bit, $\pm3\sigma$           & 8.1   & 7.9  & $-19.14$ & 61.78 & 70.92 & 4.4  & 2.32  & 82.86 & 4.51  & 7.04 \\
BAQ 3-bit, $\pm3\sigma$           & 6.1   & 10.4 & $-13.21$ & 47.40 & 55.95 & 5.5  & 0.08  & 25.00 & 0.16  & 5.09 \\
BAQ 2-bit, $\pm3\sigma$           & 4.1   & 15.5 & $-7.23$  & 21.74 & 28.92 & 9.1  & 0.00  & --    & 0.00  & 3.26 \\
BAQ 1-bit, $\pm3\sigma$           & 2.1   & 30.1 & $-0.73$  & 0.00  & 0.00  & 14.8 & 0.00  & --    & 0.00  & 2.12 \\
\midrule
BAQ 8-bit, $\pm3.75\sigma$        & 16.1  & 4.0  & $-41.01$ & 97.11 & 97.90 & 0.3  & 52.72 & 98.65 & 68.72 & 14.40 \\
BAQ 6-bit, $\pm3.25\sigma$        & 12.1  & 5.3  & $-30.19$ & 86.44 & 90.56 & 1.5  & 19.20 & 94.12 & 31.89 & 10.81 \\
BAQ 4-bit, $\pm2.5\sigma$         & 8.1   & 7.9  & $-19.60$ & 37.04 & 51.09 & 8.6  & 0.00  & --    & 0.00  & 7.50 \\
BAQ 3-bit, $\pm2.25\sigma$        & 6.1   & 10.4 & $-14.48$ & 15.89 & 25.56 & 11.8 & 0.00  & --    & 0.00  & 5.73 \\
BAQ 2-bit, $\pm2\sigma$           & 4.1   & 15.5 & $-9.41$  & 3.10  & 5.57  & 14.0 & 0.00  & --    & 0.00  & 3.94 \\
BAQ 1-bit, $\pm1.5\sigma$         & 2.1   & 30.1 & $-4.46$  & 0.00  & 0.00  & 14.8 & 0.00  & --    & 0.00  & 2.12 \\
\midrule
FFT keep 40\%                     & 13.8  & 4.6  & $-11.37$ & 51.29 & 56.70 & 3.9 & 43.12 & 46.67 & 44.82 & -- \\
FFT keep 20\%                     & 7.4   & 8.6  & $-7.83$  & 29.13 & 34.90 & 6.5 & 24.64 & 27.52 & 26.00 & -- \\
FFT keep 10\%                     & 4.2   & 15.2 & $-5.89$  & 17.76 & 22.08 & 7.9 & 14.24 & 17.80 & 15.82 & -- \\
FFT keep 5\%                      & 2.6   & 24.6 & $-4.58$  & 12.07 & 15.03 & 8.6 & 6.16  & 10.12 & 7.66  & -- \\
FFT keep 2\%                      & 1.6   & 39.0 & $-3.33$  & 9.22  & 11.11 & 8.6 & 1.92  & 5.26  & 2.81  & -- \\
FFT keep 1\%                      & 1.3   & 48.5 & $-2.63$  & 7.52  & 9.23  & 8.9 & 0.64  & 3.33  & 1.07  & -- \\
\midrule
Global KLT, $K=2048$              & 16.0  & 4.0  & $-5.73$  & 43.14 & 39.04 & 0.2  & 9.36 & 13.09 & 10.91 & -- \\
Global KLT, $K=1024$              & 8.0   & 8.0  & $-3.34$  & 22.00 & 21.41 & 2.3  & 1.12 & 2.11  & 1.46  & -- \\
Global KLT, $K=512$               & 4.0   & 16.0 & $-2.22$  & 10.68 & 12.44 & 6.2  & 0.32 & 1.15  & 0.50  & -- \\
Global KLT, $K=256$               & 2.0   & 32.0 & $-1.54$  & 4.17  & 6.14  & 10.5 & 0.00 & 0.00  & 0.00  & -- \\
Global KLT, $K=128$               & 1.0   & 64.0 & $-1.03$  & 2.60  & 4.18  & 12.2 & 0.00 & 0.00  & 0.00  & -- \\
\midrule
Block KLT, $k=128$                & 16.0  & 4.0  & $-5.39$  & 39.86 & 34.55 & 0.6  & 7.68 & 10.88 & 9.01  & -- \\
Block KLT, $k=64$                 & 8.0   & 8.0  & $-2.92$  & 14.41 & 15.89 & 5.4  & 0.88 & 2.06  & 1.23  & -- \\
Block KLT, $k=32$                 & 4.0   & 16.0 & $-1.71$  & 5.67  & 7.75  & 9.5  & 0.16 & 1.10  & 0.28  & -- \\
Block KLT, $k=16$                 & 2.0   & 32.0 & $-1.02$  & 2.18  & 3.44  & 12.7 & 0.00 & 0.00  & 0.00  & -- \\
Block KLT, $k=8$                  & 1.0   & 64.0 & $-0.55$  & 0.34  & 0.65  & 14.4 & 0.00 & 0.00  & 0.00  & -- \\
\bottomrule
\end{tabular}
\end{table*}

\emph{The autoencoder is model-limited.} At 32~b/cs, rounding the input to
FP16 gives \SI{-73.7}{dB}, while the $C=32$ autoencoder, sending the same
number of FP16 values, gives \SI{-4.7}{dB}. The transform-domain model
(Table~\ref{tab:td}) reaches only \SI{-7.7}{dB} at 64~b/cs, where an identity
map is lossless. The fair linear comparison is the $16\times16$ block KLT.
Like the autoencoder it is local, but its encoder costs a tenth as much. It
beats the autoencoder in NMSE at every rate: \SI{-5.39}{dB} against
\SI{-2.87}{dB} at 16~b/cs, \SI{-2.92}{dB} against \SI{-1.88}{dB} at 8, and
\SI{-1.71}{dB} against \SI{-1.06}{dB} at 4. It also beats the autoencoder's
all-cell retention at 8 and 16~b/cs. The global KLT, which the
autoencoder could not represent, adds only \SIrange{0.3}{0.5}{dB} over the
block KLT at 4--16~b/cs.

\emph{BAQ dominates in NMSE.} At 16~b/cs, 8-bit BAQ reaches \SI{-35.5}{dB}
with $\pm3\sigma$ clipping and \SI{-41.0}{dB} with $\pm3.75\sigma$. FFT
thresholding, with per-patch adaptive support, beats both KLTs at similar
rates (\SI{-5.89}{dB} at 4.2~b/cs against \SI{-2.22}{dB} and \SI{-1.71}{dB} at
4~b/cs), consistent with the patch-specific concentration in
Section~\ref{sec:char}.

\emph{Clipping sets both the NMSE slope and detection.} With $\pm3\sigma$
clipping, BAQ gains \SI{16.3}{dB} from 4 to 8 bits, about \SI{4.1}{dB} per
bit. Clipping error dominates at high rates. With the clipping tuned per bit
depth, the gain is \SI{21.4}{dB}, or \SI{5.35}{dB} per bit, close to the
\SI{6.02}{dB} of an unclipped high-rate quantizer. Interior CFAR detections
are strong, isolated peaks, and they are exactly what clipping removes. At 16~b/cs,
interior F1 is 16.7\% at $\pm3\sigma$ and 68.7\% at $\pm3.75\sigma$. At 8~b/cs
or less, every BAQ variant scores at most 4.5\% interior F1, both because the
NMSE-optimal clip there is narrow ($\pm2$--$2.5\sigma$) and because coarse
levels cannot represent a peak. FFT thresholding keeps exactly these peaks
and scores 16--26\% interior F1 at 4--8~b/cs. A detection-oriented coder at
low rates should therefore preserve peaks, by clipping widely or by coding a
sparse set of transform coefficients. Entropy coding the BAQ indices saves
0.2--1.7~b/cs (column $H$), so an entropy-coded learned coder must be compared
with an entropy-coded BAQ.

\subsection{Transform-Domain Rate Sweep}

Table~\ref{tab:td} sweeps the transform-domain model. It was trained for 8
epochs with the last checkpoint kept, so its values are pessimistic relative
to V7.

\begin{table}[t]
\centering
\caption{Transform-domain autoencoder sweep (8 epochs, last checkpoint)}
\label{tab:td}
\begin{tabular}{lccccc}
\toprule
$C$ (b/cs)  & 4 & 8 & 16 & 32 & 64 \\
NMSE (dB)   & $-0.65$ & $-2.31$ & $-2.72$ & $-5.31$ & $-7.71$ \\
\bottomrule
\end{tabular}
\end{table}

\subsection{Image-Domain Illustration}

\begin{figure*}[t]
\centering
\pgfplotsset{
  patchax/.style={scale only axis, width=0.21\textwidth, height=0.21\textwidth,
    enlargelimits=false, axis on top, xmin=-0.5, xmax=63.5, ymin=-0.5, ymax=63.5,
    xtick={0,20,40,60}, ytick={3,23,43,63}, yticklabels={60,40,20,0},
    tick align=outside, tick label style={font=\footnotesize},
    label style={font=\footnotesize}, title style={font=\footnotesize, yshift=-2pt}},
  cbax/.style={scale only axis, width=0.012\textwidth, height=0.21\textwidth,
    enlargelimits=false, axis on top, xmin=0, xmax=1, xtick=\empty,
    ytick pos=right, yticklabel pos=right, tick align=outside,
    tick label style={font=\footnotesize}},
}
\begin{tikzpicture}
\begin{axis}[patchax, name=a1, title={Raw: FFT magnitude (dB)},
  xlabel={Cross-range bin}, ylabel={Range bin}]
\addplot graphics[xmin=-0.5, xmax=63.5, ymin=-0.5, ymax=63.5,
  includegraphics={trim=27.36 46.08 779.76 49.68, clip}] {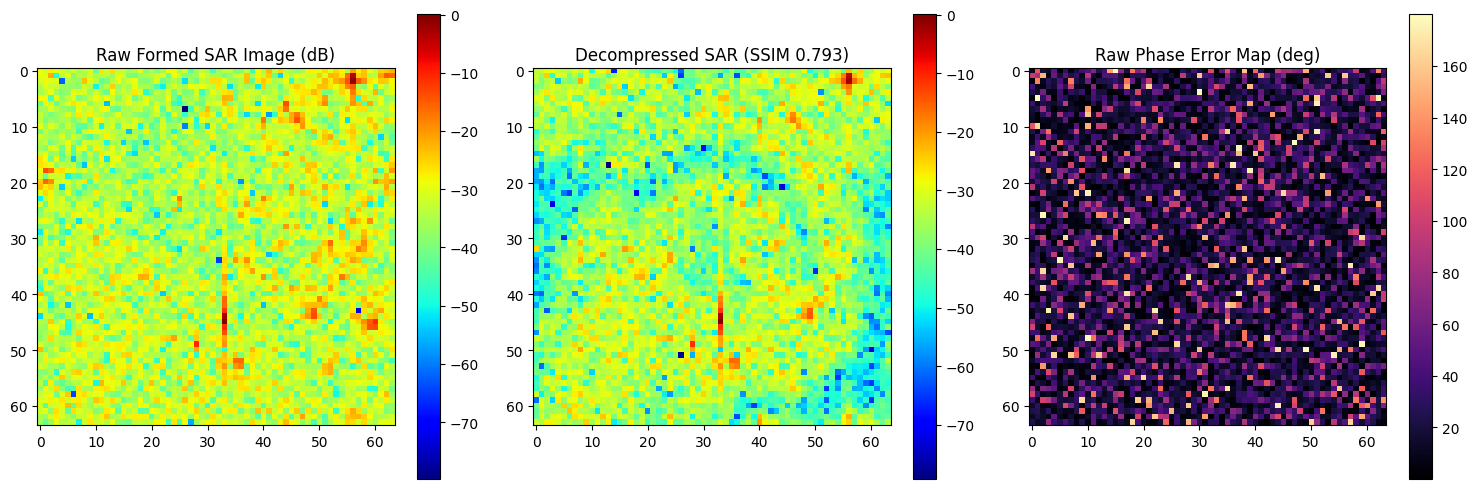};
\end{axis}
\begin{axis}[cbax, name=c1, at={(a1.south east)}, xshift=4pt, anchor=south west,
  ymin=-79.8, ymax=0, ytick={0,-20,-40,-60}]
\addplot graphics[xmin=0, xmax=1, ymin=-79.8, ymax=0,
  includegraphics={trim=300.96 7.2 747.36 10.8, clip}] {figures/sar_reconstruction.png};
\end{axis}
\begin{axis}[patchax, name=a2, at={(c1.south east)}, xshift=30pt, anchor=south west,
  title={Decoded: FFT magnitude (dB)}, xlabel={Cross-range bin}, yticklabels={}]
\addplot graphics[xmin=-0.5, xmax=63.5, ymin=-0.5, ymax=63.5,
  includegraphics={trim=384.48 46.08 422.64 49.68, clip}] {figures/sar_reconstruction.png};
\end{axis}
\begin{axis}[cbax, name=c2, at={(a2.south east)}, xshift=4pt, anchor=south west,
  ymin=-79.2, ymax=0, ytick={0,-20,-40,-60}]
\addplot graphics[xmin=0, xmax=1, ymin=-79.2, ymax=0,
  includegraphics={trim=658.08 7.2 390.24 10.8, clip}] {figures/sar_reconstruction.png};
\end{axis}
\begin{axis}[patchax, name=a3, at={(c2.south east)}, xshift=66pt, anchor=south west,
  title={Phase error (deg)}, xlabel={Pulse}, ylabel={Frequency sample}]
\addplot graphics[xmin=-0.5, xmax=63.5, ymin=-0.5, ymax=63.5,
  includegraphics={trim=741.6 46.08 66.24 49.68, clip}] {figures/sar_reconstruction.png};
\end{axis}
\begin{axis}[cbax, name=c3, at={(a3.south east)}, xshift=4pt, anchor=south west,
  ymin=0, ymax=180, ytick={0,60,120,180}]
\addplot graphics[xmin=0, xmax=1, ymin=0, ymax=180,
  includegraphics={trim=1015.2 7.2 33.12 10.8, clip}] {figures/sar_reconstruction.png};
\end{axis}
\end{tikzpicture}
\caption{One test patch at 16 b/cs (V6). Left: FFT magnitude of the raw patch,
a coarse image (dB relative to its peak). Center: the same for the decoded
patch. Right: per-sample phase error in the phase-history domain (degrees).}
\label{fig:sar}
\end{figure*}
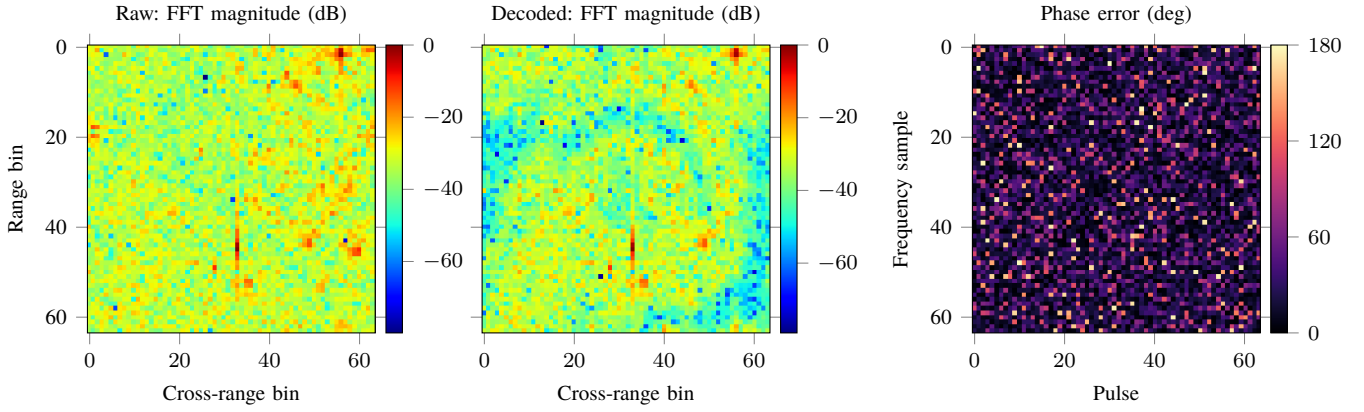

Fig.~\ref{fig:sar} shows one test patch at 16~b/cs (V6), with its FFT
magnitude as a coarse image. The dominant scatterer and a few secondary
returns survive, but the low-level background is suppressed by more than
\SI{10}{dB} in places. The focused complex coherence on this patch is 0.813.
The example is a single patch and is illustrative only.

\subsection{Computational Footprint}

Table~\ref{tab:cx} gives exact parameter and multiply-accumulate (MAC)
counts. The $C=16$ encoder needs 21.8~MMAC per patch, about \num{5300} MAC per
complex sample. The block KLT at the same rate needs 512, and BAQ needs a few
operations per sample: a block variance, a scale, and a comparison. At a
complex sample rate $f_s$, the autoencoder needs $5.3\times10^{3}f_s$~MAC/s,
which is \SI{0.53}{TMAC/s} at \SI{100}{MS/s} and \SI{5.3}{TMAC/s} at
\SI{1}{GS/s}. On-board radar processors are usually FPGAs or
radiation-tolerant devices. Sustaining this throughput there, even at INT8,
would be a substantial cost that we have not demonstrated. We have no
on-target latency or power measurements.

\begin{table}[t]
\centering
\caption{Exact model complexity per $64\times64$ patch (conv/deconv MACs)}
\label{tab:cx}
\begin{tabular}{rrrrr}
\toprule
 & \multicolumn{2}{c}{Encoder} & \multicolumn{2}{c}{Decoder} \\
\cmidrule(lr){2-3}\cmidrule(lr){4-5}
$C$ (b/cs) & Params & MMAC & Params & MMAC \\
\midrule
4  & \num{20772}  & 19.83 & \num{22106}  & 20.52 \\
8  & \num{22824}  & 20.35 & \num{25042}  & 21.27 \\
16 & \num{28656}  & 21.82 & \num{32642}  & 23.20 \\
32 & \num{47232}  & 26.54 & \num{54754}  & 28.84 \\
\bottomrule
\end{tabular}
\end{table}

\section{Discussion}
\label{sec:disc}

\subsection{What Limits the Autoencoder}

The evidence supports a narrow conclusion. \emph{This} 28k-parameter
encoder, trained for 15 epochs with a non-MSE loss on globally normalized
data, loses to BAQ, and also to a local linear transform with a tenth of its
encoder cost. It does not show that learned coding of phase history is
fundamentally limited. Three factors are confounded with ``learned versus
not'':
\begin{itemize}
  \item \emph{Capacity and optimization.} The network cannot approach an
        identity map at 32--64~b/cs, and it underperforms the block KLT at
        matched rates.
  \item \emph{Normalization.} BAQ normalizes each block, and within a patch
        the block scales differ by only \SI{1.9}{dB}, so this is essentially
        per-patch gain normalization. The network instead sees patches whose
        RMS level spans \SI{16}{dB} and sits 20--\SI{36}{dB} below full scale.
        Batch normalization in the decoder path does not undo a
        per-sample gain.
  \item \emph{Objective.} The loss (\ref{eq:loss}) is not MSE, so NMSE is
        not what the model optimizes.
\end{itemize}
The data do constrain what any coder can gain. Local correlation is modest
(Table~\ref{tab:char}), and the spectral concentration is patch-specific,
which favors per-patch adaptive methods. The most informative next
experiment separates normalization from learning. Dividing each patch by its
RMS and sending one FP16 scale costs $16/4096\approx0.004$~b/cs. A network
trained on such input, or one that learns a residual on top of BAQ, would
test whether learning adds anything once the gain confound is removed.

\subsection{Relation to Image Formation}

Since the data are deramped k-space samples (Section~\ref{sec:raw}), the 2-D
FFT of a patch is close to a coarse sub-aperture image. Much of the Fourier
concentration in Table~\ref{tab:char} reflects bright scatterers in the
scene. V7, which operates on this spectrum, is close to compressing a coarse
image. This fits the view that learned compression pays off for SAR mainly
after image formation, where focused imagery~\cite{asiyabi2023complex} has
the spatial structure that learned codecs exploit.

\subsection{What the CFAR Proxy Measures}
\label{sec:cfar}

On raw test patches, CA-CFAR fires on \num{41774} of $2.83\times10^{7}$
cells, an empirical $P_{fa}$ of $1.5\times10^{-3}$. Of these detections, 97\%
lie in the 6-cell border band. There the zero-padded training window
underestimates the local mean, and the empirical rate is $4.2\times10^{-3}$,
$42\times$ the design value. On the interior cells it is $6.7\times10^{-5}$,
below the design $P_{fa}$, with \num{1250} detections. All-cell scores
(retention, all-cell F1, and every autoencoder detection number in this
paper) therefore mostly measure how a coder perturbs the energy near patch
borders. Interior scores measure how well it preserves strong, isolated peaks
in the phase-history power map. The two can rank methods differently.
$\pm3\sigma$ BAQ at 8 bits has 87\% all-cell F1 but 17\% interior F1, while
FFT thresholding keeps 45\% interior F1 at 13.8~b/cs. Neither is target
detection. A meaningful operational test forms full-aperture images from the
decompressed phase history and scores detection of the known GOTCHA targets
and calibration reflectors.

\subsection{A Recommended Evaluation Protocol}
\label{sec:protocol}

Each item below guards against an error that makes a learned coder look
better than it is.
\begin{enumerate}
  \item \emph{Derive the rate from the bits sent.} Take the latent shape and
        dtype from the encoder output, and report bits per complex sample. A
        stale latent size overstated our ratio $16\times$, and a spurious
        ``complex'' factor understated it $2\times$.
  \item \emph{Charge all side information.} FFT thresholding credited with
        $10{:}1$ at FP32 and with free positions is really $15.2{:}1$ at
        4.2~b/cs.
  \item \emph{Use real baselines at their best settings.} Use block-adaptive
        quantization with tuned clipping and an entropy-coded variant, not a
        global uniform quantizer (which reaches only \SI{-1.5}{dB} at 4 bits,
        against \SI{-19.6}{dB}). Add an FP16 cast and a block KLT as sanity
        anchors.
  \item \emph{Evaluate every method on the same full test set,} not on one
        batch.
  \item \emph{Estimate coherence per patch.} Pooled estimators cancel
        patch-specific phase and can suggest that the data are white.
  \item \emph{Score detection one-to-one, away from borders.} Report recall,
        precision, and $\Delta P_{fa}$ on cells where the CFAR window is
        complete.
  \item \emph{Report seed spread.} Runs differing only in seeding and
        checkpoint selection spanned \SI{2.5}{dB} NMSE, and even one seed
        varied by \SI{0.4}{dB} across runs.
  \item \emph{Split by acquisition,} by pass or azimuth sector, not by file.
\end{enumerate}

\section{Limitations and Future Work}

The study uses one dataset and fixed $64\times64$ patches of deramped k-space
data, not ADC samples. The autoencoders were trained on a file-level split
that leaks pass and azimuth geometry into the test set, and they were not
retrained on a held-out split. For the linear coders this leakage is worth at
most \SI{0.16}{dB}. Training runs are short, seeds were varied at one rate
only, and latents are unquantized. Autoencoder checkpoints were not retained,
so their detection scores are all-cell bounds that cannot be recomputed on
interior cells. There are no on-target hardware measurements. In priority
order, the next steps are:
\begin{enumerate}
  \item a per-patch-normalized autoencoder, with the scale sent as side
        information, or a learned residual on top of BAQ;
  \item a retrained autoencoder sweep scored on interior cells;
  \item quantized, entropy-coded latents trained under a rate-distortion
        objective~\cite{balle2018variational}, compared against entropy-coded
        BAQ;
  \item a pass-held-out split for the learned models;
  \item detection on formed full-aperture imagery with ground-truth targets;
  \item throughput and power on a representative FPGA or radiation-tolerant
        target.
\end{enumerate}

\section{Conclusion}

We benchmarked a lightweight split autoencoder for on-board compression of
GOTCHA phase history, charging every method for every transmitted bit and
reporting rates in bits per complex sample. The autoencoder lost to BAQ at
every rate, by \SI{32.6}{dB} NMSE at 16~b/cs, or \SI{38.1}{dB} against BAQ
with tuned clipping. It also lost to a $16\times16$ block KLT with a tenth of
its encoder cost. The evidence attributes the loss mainly to the model, its
normalization, and its objective, not to a fundamental property of the data.
Two results concern evaluation more broadly. CFAR scores on small raw patches
are dominated by border artifacts unless restricted to interior cells. On
those cells, the clipping range, not the bit depth alone, decides whether BAQ
preserves detections, and at low rates sparse transform coding preserves
them better. The first experiment to run next is a learned coder on top of
BAQ's per-patch normalization.

\section*{Acknowledgment}
The phase-history data used in this work are the Gotcha Volumetric SAR Data
Set, Version~1.0, provided by the Air Force Research Laboratory, Sensors
Directorate (AFRL/SNA), which is gratefully acknowledged as the source of the
data (public release no.\ SN-07-0045, AFRL/WS-07-0528).


\end{document}